\documentclass[letterpaper]{article} 
\usepackage[submission]{aaai2026}  
\usepackage{times}  
\usepackage{helvet}  
\usepackage{courier}  
\usepackage[hyphens]{url}  
\usepackage{graphicx} 
\usepackage{natbib}  
\usepackage{caption} 
\usepackage{booktabs} 
\usepackage{algorithm}
\usepackage{algorithmic}
\usepackage{multirow} 
\usepackage{array}

\newcolumntype{C}{>{\centering\arraybackslash}p{1.2cm}}

\usepackage{newfloat}
\usepackage{listings}
\DeclareCaptionStyle{ruled}{labelfont=normalfont,labelsep=colon,strut=off} 
\floatstyle{ruled}
\newfloat{listing}{tb}{lst}{}
\floatname{listing}{Listing}
\title{Reconstruction and Reenactment Separated Method for Realistic Gaussian Head}
\author {
    Zhiling Ye,
    Cong Zhou,
    Xiubao Zhang,
    Haifeng Shen,
    Weihong Deng,
    Quan Lu
}
\affiliations{
    Mashang Consumer Finance Co., Ltd.\\
   \{zhiling.ye,  cong.zhou01, xiubao.zhang, haifeng.shen, weihong.deng, quan.lu\}@msxf.com

}
\usepackage{bibentry}

\begin{document}

\maketitle


\begin{abstract}

In this paper, we explore a reconstruction and reenactment separated framework for 3D Gaussians head, which requires only a single portrait image as input to generate controllable avatar. Specifically, we developed a large-scale one-shot gaussian head generator built upon WebSSL \cite{fan2025webssl} and employed a two-stage training approach that significantly enhances the capabilities of generalization and high-frequency texture reconstruction. During inference, an ultra-lightweight gaussian avatar  driven by control signals enables high frame-rate rendering, achieving 90 FPS at a resolution of \(512\times512\). We further demonstrate that the proposed framework follows the scaling law, whereby increasing the parameter scale of the reconstruction module leads to improved performance. Moreover, thanks to the separation design, driving efficiency remains unaffected. Finally, extensive quantitative and qualitative experiments validate that our approach outperforms current state-of-the-art methods.

\end{abstract} 

\begin{figure*}
    \vspace{-1cm}
    \includegraphics[width=1\linewidth]{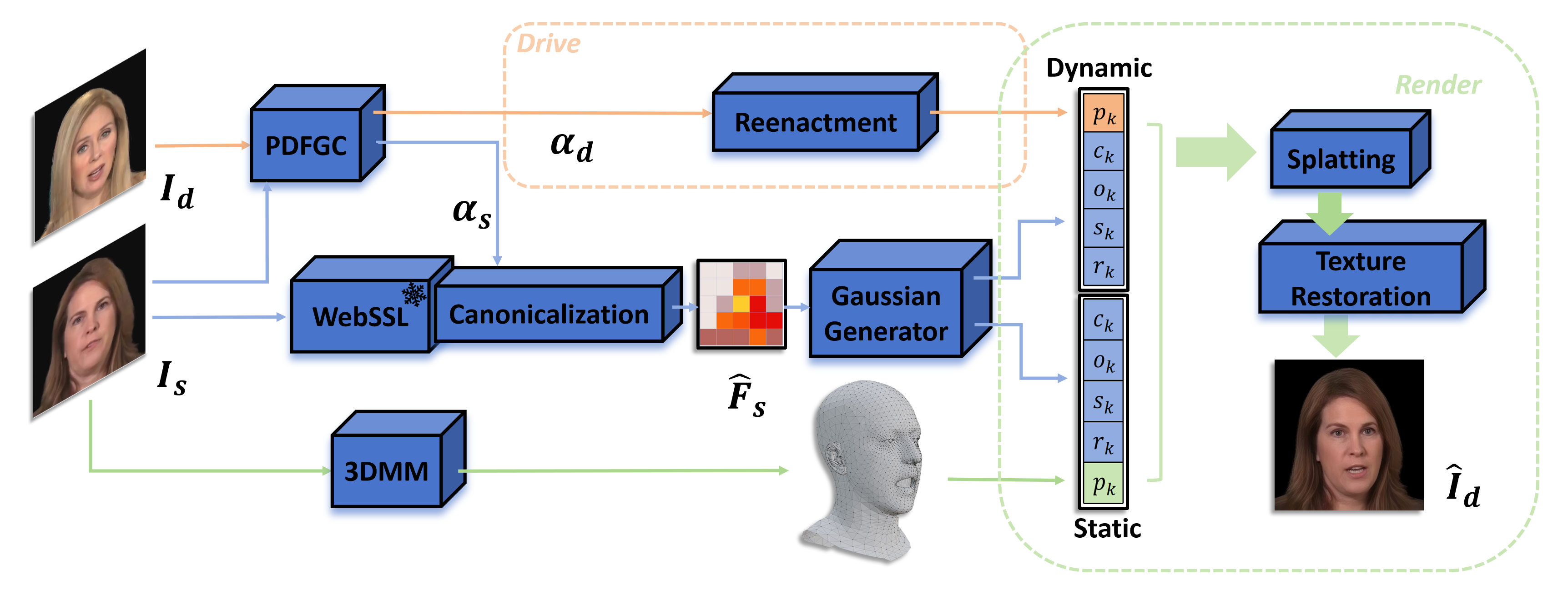}
    \vspace{-0.5cm}
    \caption{\textbf{The full pipeline of RAR.} For source portrait image \(I_s\), we extract control condition \(\alpha_{d}\) from driving image \(I_d\) to independently adjust features (eyes, mouth, expressions) and obtain the rendered \(\hat{I_d}\). Firstly, we extract the corresponding canonical feature \(\hat{F_s}\) from \(I_s\). Then the gaussian generator predicts appearance details (colors, opacity, scales), splitting them into static \(\mathcal{G}_{static}\) and dynamic \(\mathcal{G}_{dynamic}\) parts. The positions of static part \(\mathcal{G}_{static}\) were proposed by a mesh of 3D Morphable Models, namely FLAME \cite{li2017learning}. The reenactment module maps \(\alpha_{d}\) to the positions \( p_k \) of dynamic part \(\mathcal{G}_{dynamic}\). The image splatted from static and dynamic 3D Gaussians was refined by the texture restoration module to produce the final output  \(\hat{I_d}\). More details about each module can be found in the supplementary material.}
    \label{fig:full_pipe}
    \vspace{-0.6cm}
\end{figure*}
\section{Introduction}


Avatar reconstruction from a single portrait image is challenging yet promising with applications in video conferencing, filmmaking, and game production. Researchers globally have explored solutions using end-to-end 2D methods and approaches that leverage 3D priors.

Specifically, 2D-based approaches \cite{goodfellow2014generative, isola2017image, karras2019style, karras2020stylegan, guo2024liveportrait} mainly rely on deep convolutional networks and generative adversarial models, achieving good control of facial expressions and body postures in the source image through the construction and modulation of warp field. Notably, in recent years, with the rapid development of generative models such as image and video diffusion models, research based on 2D schemes has achieved a series of significant advancements \cite{cui2024hallo2,tian2024emo,chen2025hunyuanvideo}. However, the distinct advantages were demonstrated in many practical applications, their inherent limitation lies in the lack of explicit 3D structural priors. The lack of 3D priors as guidance leads to the need of more complex model structures and larger model sizes for 2D methods to achieve end-to-end solutions. Consequently, this not only requires substantial computational resources but also inevitably introduces higher latency issues.

On the other hand, cutting-edge 3D synthesis technologies such as NeRF (Neural Radiance Fields) \cite{mildenhall2020nerf} and 3DGS (3D Gaussian Splatting) \cite{kerbl2023gaussian} can be employed to achieve efficient and accurate character avatar reconstruction and reenactment, while effectively maintaining consistency and coherence from multiple viewpoints. It is worth noting that although these methods \cite{gafni2021dynamic,bai2023high,yu2023nofa,zheng2023pointavatar,chu2024gpaavatar,chu2024generalizable,he2025lam} have demonstrated high precision in both theoretical and experimental settings, they generally rely on the precise estimation of the 3D pose of the character from a single image. This step introduces estimation errors that can cause texture inaccuracies and expression distortions. Additionally, the methods rely heavily on the 3DMM mesh, which is a
widely-used 3D morphable model, to drive expressions, but its limited capabilities make reproducing subtle, natural expressions challenging.

We proposed a \textbf{R}econstruction \textbf{A}nd \textbf{R}eenactment separated method, named \textbf{RAR}. RAR predicts 3D Gaussians from a single head image, allowing control over eyes, mouth shape, and expression. Our decoupled architecture separates appearance reconstruction and expression control, ensuring precise reconstruction and high frame-rate reenactment. The appearance feature is extracted by using WebSSL-7B \cite{fan2025webssl} to create a standardized feature map with a canonical expression, which is then processed by a 3D Gaussian Generator to obtain texture and structural details. A lightweight reenactment module converts driving information into positions that reenact the 3D gaussian head at 90 FPS. Our method outperforms most state-of-the-art approaches on extensive experiments and comparisons.

And errors introduced by 3D estimation can result in inaccurate recovery of appearance textures during avatar reconstruction. Such artifacts are particularly prone to occur in regions with high-frequency textures, such as teeth, hair, and eyelashes. On the other hand, 2D end-to-end approaches \cite{wang2021oneshot, guo2024liveportrait} use a warping field to distort the input image and generate pseudo-3D motions. These distortions approximate 3D motion using relatively simple transformations in 2D space, making it difficult for the human visual system to distinguish between these 2D transformations and true 3D motion. However, in terms of algorithmic learning complexity, the pseudo-3D motions patterns generated by the 2D end-to-end approach are much easier to learn than real 3D motion patterns. Therefore, to eliminate the artifacts induced by errors in 3D estimation, we add an additional texture restoration module and synthetic data in a second phase after the pre-training. Specifically, we freeze the previously trained model and train the texture restoration module solely using synthetic data. In this work, we use the 2D end-to-end model Live Portrait \cite{guo2024liveportrait} to generate the synthetic data. To address the challenge of capturing fine-grained expression details. We abandon control schemes based on 3DMM  \cite{li2017learning} or geometric structures and introduce a approach based on implicit parameters. This allows for independent control of facial features such as the eyes and mouth, and it can reproduce delicate and natural expressions. Leveraging this characteristic, we train an audio-to-motion (audio2motion) model that maps speech to mouth shape parameters, achieving a speech-driven avatar reconstruction and further validating the excellent control capability. This speech-driven experiment can can be found in the supplementary material.

Our main contributions are as follows: 
\begin{itemize}
    \item  We designed a decoupled architecture for appearance reconstruction and expression reenactment, employing a larger-scale WebSSL backbone to enhance the generalization and accuracy of the reconstruction module. meanwhile, the expression driving utilizes a lightweight model to achieve a driving speed of 90 FPS.  
    \item  We introduced a texture restoration module and a two-stage training process using synthetic data, effectively eliminating artifacts in high-frequency textures caused by 3D reconstruction errors.
    \item Extensive experiments and comparisons on a benchmark dataset verify the effectiveness and breakthrough performance of our method.
    \item Finally, our experiments show that the separation architecture follows the scaling law. Namely increasing the parameter scale of the reconstruction module improves the performance of algorithm. This enhancement does not compromise efficiency during driving.
\end{itemize}

\begin{figure*}[t]
\vspace{-0.8cm}
\centering
\includegraphics[width=2\columnwidth]{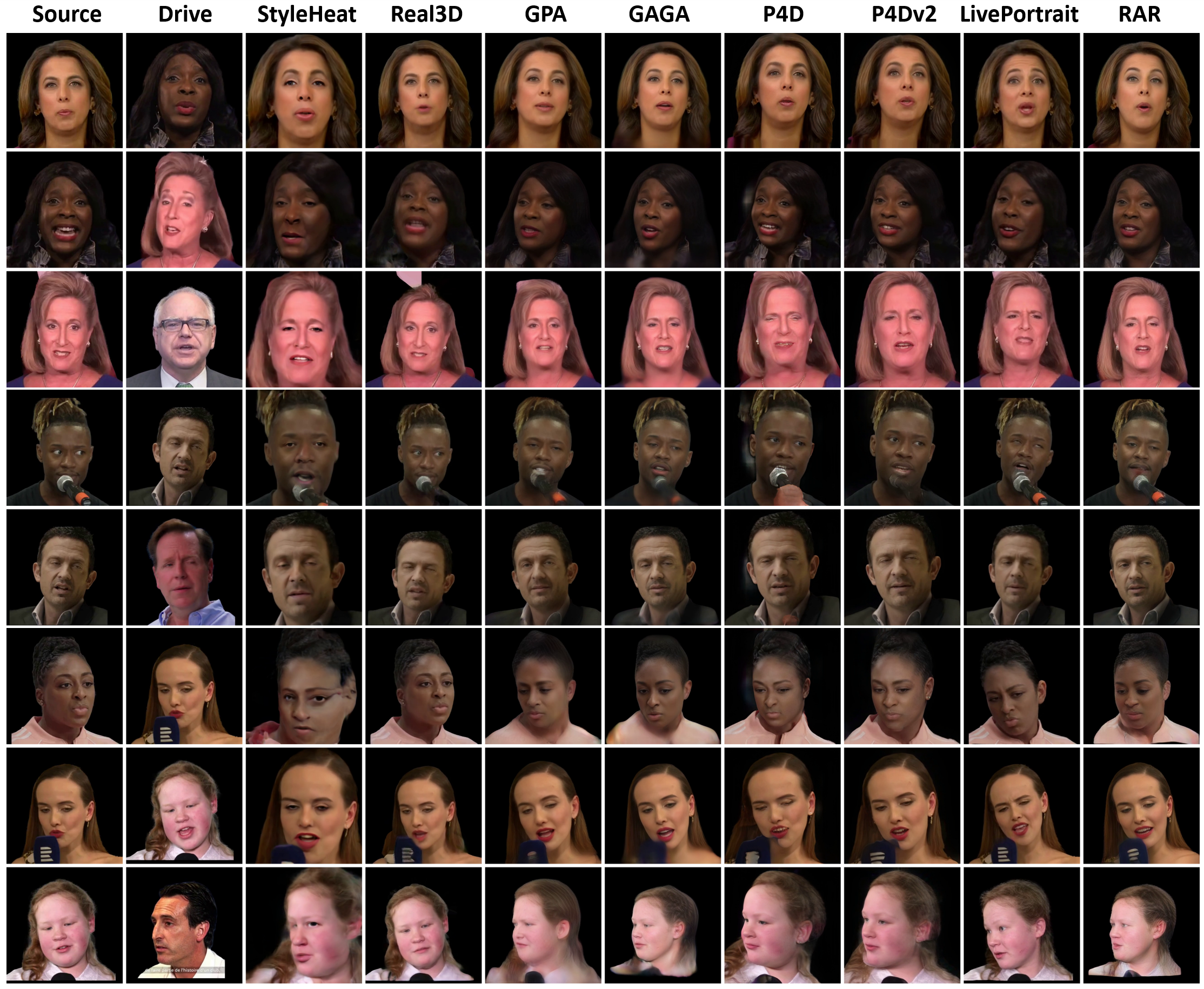} 
\caption{Visualization of cross-reenacted results on the VFHQ and HDTF dataset.}
\label{fig:cross reenactment}
\vspace{-0.3cm}
\end{figure*}
\section{Related work}
Recent advances in generating controllable head avatar with single portrait image can be broadly divided into two categories: 2D end-to-end image synthesis approaches and 3D explicit structural prior-based methods.

The 2D techniques primarily utilize convolutional neural networks (CNNs) \cite{goodfellow2014generative, isola2017image, karras2020stylegan} to achieve end-to-end image synthesis, and they extensively employ generative adversarial networks (GANs). Early methods \cite{zakharov2019fewshot, burkov2020neural, wang2023progressive} focused on integrating both expression and pose features into the generator network, typically using architectures such as U-Net or StyleGAN. Other approaches in this category \cite{burkov2020neural,  guo2024liveportrait, hong2022depth, zhang2023metaportrait} represent expressions and poses as deformation fields applied to the source image. Thanks to advancements in image and video diffusion networks, recently diffusion model-based techniques \cite{cui2024hallo2, tian2024emo, chen2025hunyuanvideo} have been adopted to improve synthesis quality. However, these methods still face challenges such as high latency and computational resource demands. Moreover, the lack of an explicit 3D structure in 2D-based approaches often leads to unrealistic distortions or changes in identity features when handling significant pose or expression variations. Although some methods \cite{blanz1999morphable, gerig2018morphable, li2017learning, paysan20093dface} introduce 3D morphable model, namely 3DMM, \cite{li2017learning} to mitigate these issues, they ultimately fall short in supporting free-viewpoint rendering due to insufficient concrete 3D structural constraints.
In contrast, methods incorporating 3D structural priors offer stronger geometric consistency and free-viewpoint rendering capabilities. Early 3D approaches \cite{khakhulin2022realistic, xu2020deep3d} used 3DMM for head avatar reconstruction, and later, the emergence of neural radiance fields (NeRFs) \cite{mildenhall2020nerf} prompted numerous recent methods \cite{bai2023high, chu2024gpaavatar, deng2024portrait4d, deng2024portrait4dv2, chu2024generalizable, he2025lam, ye2024real3d, yu2023nofa, zheng2023pointavatar, zielonka2023instant, ye2025realistic}. However, NeRF-based methods typically require large amounts of training data, including multi-view or even monocular videos, which raises privacy concerns and limits their generalization to unseen identities. Some methods \cite{sun2023next3d, sun2024cgof, tang20243dfaceshop,xu2023pv3d, zhuang2022controllable} attempt to bypass the need for extensive datasets by training the generator with random noise and then reconstructing specific identities through inversion, though the accuracy of inversion remains challenging. Additionally, test-time optimization has been explored as an alternative, but its computational cost hinders practical application.

Recent studies \cite{chu2024gpaavatar, deng2024portrait4d, deng2024portrait4dv2, li2023generalizable, han2024cvthead, ye2024real3d} have investigated single-sample 3D head reconstruction methods to address the limitations in data requirements and computational efficiency. These methods leverage various techniques, including tri-plane features, deformation fields, point-based expression fields, and vertex feature transformers. Despite these advances, NeRF-based approaches still struggle to achieve real-time rendering. Recently, 3D Gaussian Splatting \cite{kerbl2023gaussian} has emerged as a promising alternative, capable of generating high-quality results at significantly faster rendering speeds. However, existing gaussian splatting methods \cite{qian2024gaussianavatars, xu2024gaussianhead} typically depend on training with video data of specific individuals, thus limiting their generalization to novel identities. The latest work, GAGAvatar \cite{chu2024generalizable} and LAM \cite{he2025lam}, proposes a single-sample 3D Gaussian head digital avatar generation method. however, it still relies on precise 3D pose estimation of the training data. The inherent errors in estimating 3D poses from monocular videos make it challenging to accurately reconstruct the appearance of subject.

Our method not only makes effective use of 3D structural priors to maintain geometric consistency, free-viewpoint rendering capability, and efficiency, but also leverages the high visual quality of 2D end-to-end synthesis to address the high-frequency texture degradation caused by pose estimation errors in 3D methods.

\begin{table*}[htbp]
\vspace{-0.8cm}
\small
\setlength{\tabcolsep}{2.5pt}
\renewcommand{\arraystretch}{1.1}
\centering


\hfill
\begin{minipage}{5\linewidth}
\begin{tabular}{l | c c c c c c c c | c c c}

\toprule
\hline
\multirow{2}{*}{\textbf{Method}} & 
\multicolumn{8}{c}{\textbf{Self Reenactment}} & 
\multicolumn{3}{|c}{\textbf{Cross Reenactment}} \\
\cmidrule(l){2-9} 
\cmidrule(l){10-12}
&
\rotatebox[origin=c]{0}{\textbf{CSIM↑}} & 
\rotatebox[origin=c]{0}{\textbf{LPIPS↓}} & 
\rotatebox[origin=c]{0}{\textbf{SSIM↑}} & 
\rotatebox[origin=c]{0}{\textbf{PSNR↑}} & 
\rotatebox[origin=c]{0}{\textbf{FLMD↓}} & 
\rotatebox[origin=c]{0}{\textbf{MLMD↓}} & 
\rotatebox[origin=c]{0}{\textbf{APC↑}} & 
\rotatebox[origin=c]{0}{\textbf{AEC↑}} & 
\rotatebox[origin=c]{0}{\textbf{CSIM↑}} & 
\rotatebox[origin=c]{0}{\textbf{APC↑}} &
\rotatebox[origin=c]{0}{\textbf{AEC↑}} \\
\hline
\textbf{P4D \cite{deng2024portrait4d}} 
&0.704	&0.286	&0.676	&16.688	&2.661	&3.806	&0.843	&0.836	&0.361	&0.586	&0.326 \\
\textbf{P4D-v2 \cite{deng2024portrait4dv2}} 
&0.744 &0.262 &0.691 &17.272 &2.408 &3.529 &0.910 &0.897 &0.379 &0.644 &0.383 \\
\textbf{GPAvatar \cite{chu2024gpaavatar}} 
&0.737 &0.246 &0.718 &18.257 &2.223 &2.869 &0.852 &0.835 &0.386 &0.586 &0.418 \\
\textbf{LivePortrait \cite{guo2024liveportrait}} 
&0.808 &0.212 &0.733 &19.386 &1.806 &2.611 &0.966 &\underline{0.944} &0.381 &0.664 &0.410 \\
\textbf{StyleHEAT  \cite{yin2022styleheat}} 
&0.626 &0.337 &0.655 &15.439 &2.827 &5.071 &0.819 &0.789 &0.313 &0.668 &0.372 \\
\textbf{Real3DPortrait \cite{ye2024real3d}} 
&0.750 &0.251 &0.726 &18.598 &1.967 &3.031 &0.938 &0.875 &0.272 &0.634 &0.378 \\
\textbf{GAGAvatar  \cite{chu2024generalizable}} 
&0.826 &0.207 &0.746 &19.473 &1.315 &1.921 &0.963 &0.940 &0.405 &0.713 &0.450 \\
\hline
\textbf{RAR} 
& \textbf{0.836} & \underline{0.181} & \underline{0.763} &\underline{19.738} & \underline{1.226} & \underline{1.788} & \underline{0.973} & \underline{0.944} & \textbf{0.425} & \underline{0.738} & \underline{0.471} \\
\textbf{RAR (fine-tune)} 
& \underline{0.832} & \textbf{0.180} & \textbf{0.773} & \textbf{19.832} & \textbf{1.214} & \textbf{1.709} & \textbf{0.985} & \textbf{0.950} & \underline{0.415} & \textbf{0.748} & \textbf{0.487} \\
\bottomrule
\end{tabular}
\end{minipage}
\vspace{0.1cm}

\caption{Quantitative comparison on the VFHQ dataset. We highlight the \textbf{best} and \underline{second best} results.} 
\label{tab:comparsion_VFHQ}
\vspace{-0.1cm}
\end{table*}

\begin{table*}[htbp]
\small
\setlength{\tabcolsep}{2.5pt}
\renewcommand{\arraystretch}{1.1}
\centering

\hfill
\begin{minipage}{5\linewidth}
\begin{tabular}{l | c c c c c c c c | c c c}

\toprule
\hline
\multirow{2}{*}{\textbf{Method}} & 
\multicolumn{8}{c}{\textbf{Self Reenactment}} & 
\multicolumn{3}{|c}{\textbf{Cross Reenactment}} \\
\cmidrule(l){2-9} 
\cmidrule(l){10-12}
&
\rotatebox[origin=c]{0}{\textbf{CSIM↑}} & 
\rotatebox[origin=c]{0}{\textbf{LPIPS↓}} & 
\rotatebox[origin=c]{0}{\textbf{SSIM↑}} & 
\rotatebox[origin=c]{0}{\textbf{PSNR↑}} & 
\rotatebox[origin=c]{0}{\textbf{FLMD↓}} & 
\rotatebox[origin=c]{0}{\textbf{MLMD↓}} & 
\rotatebox[origin=c]{0}{\textbf{APC↑}} & 
\rotatebox[origin=c]{0}{\textbf{AEC↑}} & 
\rotatebox[origin=c]{0}{\textbf{CSIM↑}} & 
\rotatebox[origin=c]{0}{\textbf{APC↑}} &
\rotatebox[origin=c]{0}{\textbf{AEC↑}} \\
\hline
\textbf{P4D \cite{deng2024portrait4d}} 
&0.797 &0.206 &0.729 &18.687 &1.546 &2.183 &0.917 &0.841 &0.399 &0.757 &0.348 \\
\textbf{P4D-v2 \cite{deng2024portrait4dv2}} 
&0.841 &0.187 &0.740 &18.932 &1.283 &1.903 &0.959 &0.906 &0.432 &0.791 &0.379 \\
\textbf{GPAvatar \cite{chu2024gpaavatar}} 
&0.860 &0.164 &0.781 &20.797 &0.993 &1.514 &0.963 &0.892 &0.438 &0.782 &0.397 \\
\textbf{LivePortrait \cite{guo2024liveportrait}} 
&0.882 &0.154 &0.780 &21.042 &1.005 &1.406 &0.980 &0.951 &0.438 &0.786 &0.374 \\
\textbf{StyleHEAT  \cite{yin2022styleheat}} 
&0.787 &0.264 &0.683 &15.923 &1.449 &2.416 &0.952 &0.851 &0.424 &0.786 &0.392 \\
\textbf{Real3DPortrait \cite{ye2024real3d}} 
&0.839 &0.183 &0.766 &20.263 &1.139 &1.679 &0.962 &0.902 &0.379 &0.771 &0.355 \\
\textbf{GAGAvatar  \cite{chu2024generalizable}} 
&0.878 &0.163 &0.783 &20.822 &0.848 &1.165 &0.979 &0.933 &0.451 &0.807 &0.400 \\
\hline
\textbf{RAR} 
&\underline{0.884}	&\underline{0.146}	&\underline{0.785}	&\underline{21.084}	&\underline{0.811}	&\underline{1.172}	&\underline{0.987}	&\underline{0.964}	&\textbf{0.466}	&\underline{0.817}	&\underline{0.407} \\
\textbf{RAR (fine-tune)} 
&\textbf{0.885}	&\textbf{0.142}	&\textbf{0.789}	&\textbf{21.107}	&\textbf{0.808}	&\textbf{1.153}	&\textbf{0.989}	&\textbf{0.977}	&\underline{0.460}	&\textbf{0.826}	&\textbf{0.427} \\
\bottomrule
\end{tabular}
\end{minipage}
\vspace{0.1cm}
\caption{Quantitative comparison on the HDTF dataset. We highlight the \textbf{best} and \underline{second best} results.} 
\label{tab:comparsion_HDTF}
\vspace{-0.4cm}
\end{table*}

\section{Methodology}
As illustrated in the Fig.\ref{fig:full_pipe}, RAR is a reconstruction and reenactment separated scheme that predicts the corresponding 3D Gaussian representation \(\mathcal{G}\), with the single input portrait image \(I_s\).  By utilizing the driving image \(I_d\), we extract an implicit control condition \(\alpha_{d}\), which enables independent manipulation of the eyes, mouth, and expressions to achieve the desired rendered result \(\hat{I_d}\).  For the source image \(I_s\), our reconstruction module predicts appearance-related information including colors, opacity, and scales of the portrait. This appearance-related information is split into static \(\mathcal{G}_{static}\) and dynamic \(\mathcal{G}_{dynamic}\) parts. The positions of static \(\mathcal{G}_{static}\) is proposed by FLAME which is a widely-used 3DMM model. On the other hand, the reenactment module maps the implicit control parameters \(\alpha_{d}\) to positions \( p_k \) of dynamic part \(\mathcal{G}_{dynamic}\). These appearance-related attributes and position-related attributes are integrated into a complete 3D Gaussian representation, which is subsequently refined by the texture restoration  module to produce the final output.


\subsection{Condition Extraction}
We estimate a shape mesh (excluding expressions and poses) from the source image \(I_s\) using the FLAME \cite{li2017learning} model. The positions of 5,023 vertices in this mesh serve as the fixed points for the static component of the 3D Gaussian representation, denoted as \(\{p_k\}_{static} \). The corresponding appearance attributes (colors, etc.) are predicted by the gaussian generator which we will introduced in the following part. 

PDFGC \cite{wang2023progressive} decomposes facial dynamics (e.g., lip motion, head pose, eye gaze/blink) into orthogonal latent codes through progressive representation learning. Our framework adopts PDFGC to extract control priors from images. For both the source image \(I_s\) and driving image \(I_d\), we derive their respective control parameters \(\alpha_s\) and \(\alpha_d\) via PDFGC:

\begin{equation} 
\alpha_s = PDFGC(I_s),
\alpha_d = PDFGC(I_d),
\end{equation} 

\subsection{Reconstruction Module}
The reconstruction module comprises tow components: the feature extraction block, the canonicalization block. Differ from previous method \cite{deng2024portrait4dv2,chu2024generalizable, he2025lam}, we adopt the pre-trained WebSSL as the feature extraction block. Compared with DinoV2 \cite{oquab2024dinov2}, WebSSL benefits from a larger scale and a more extensive training dataset, thereby enabling the extraction of superior features for more effective downstream task performance. A input portrait image \(I_s\) is passed through the feature extraction block to obtain its corresponding feature representation \(F_s\). The feature representation  \(F_s\) of the source image along with its corresponding \(\alpha_s\) is fed into the canonicalization block for feature normalization. The canonicalization block employs cross-attention to facilitate the interaction between \(\alpha_s\)  and \(F_s\), ultimately outputting the normalized feature representation \(\hat{F_s}\). The following equations show the simplified pipeline:

\begin{equation} 
F_s = \mathcal{F}(I_s),
\hat{F_s} = \mathcal{C}(F_s, \alpha_s),
\end{equation} 
where \(\mathcal{F}\) is denoted as WebSSL-based feature extraction block, and canonicalization block is denoted as \(\mathcal{C}\). 


\subsection{Gaussian Generator}
The Gaussian Generator be made up of 2D convolutional layers and is divided into into two components. The first is Static Block which predicts static appearance-related attributes from \(\hat{F_s}\) while contains the positions derived from FLAME mesh as \( \{p_k\}_{static} \) to integrate the static 3D Gaussian. The static 3D Gaussian ensure the foundational identity representation, primarily encoding low-frequency textures (e.g., facial contours and skin tone uniformity) in rendered outputs. And Dynamic Block processes the canonical features \(\hat{F_s}\) to predict dynamic appearance attributes. These attributes, combined with PDFGC-driven positions \( \{p_k\}_{dynamic} \), form the dynamic 3D Gaussian. This component specializes in rendering high-frequency details (e.g., eyes, wrinkles, teeth) and motion dynamics (e.g., blinking, opening mouth, expression changes). The static Gaussian \(\mathcal{G}_{static}\) and dynamic Gaussian \(\mathcal{G}_{dynamic}\)  components are concatenated into a unified 3D representation \(\mathcal{G}\). Through rasterization, this hybrid representation is rendered into the final output image \(\hat{I_d}\) achieving photorealistic animation with preserved identity consistency and dynamic details. 

\begin{equation} 
\{c_k,o_k,s_k, r_k, p_k\}_{static} = \mathcal{B}_{static}(\hat{F_s}) + \{p_k\}_{static},
\end{equation} 
where \( \{c_k,o_k,s_k, r_k, p_k\}_{static}=\mathcal{G}_{static} \) is the static 3D Gaussian, and \( \mathcal{B}_{static} \)  is Static Block of Gaussian Generator.  \( \{c_k\} \), \( \{o_k\} \), \( \{s_k\} \), \( \{r_k\} \), \( \{p_k\} \) are  colors, opacity, scales, rotation and positions of 3D Gaussian respectively. For the dynamic 3D Gaussian component, we formulated as:
\begin{equation} 
\{c_k,o_k,s_k, r_k, p_k\}_{dynamic} = \mathcal{B}_{dynamic}(\hat{F_s}) + \mathcal{D}(\alpha_d),
\end{equation} 
where \( \{c_k,o_k,s_k, r_k, p_k\}_{dynamic}=\mathcal{G}_{dynamic} \) is the dynamic 3D Gaussian. \( \mathcal{B}_{dynamic} \)  is Dynamic Block of Gaussian Generator, and \( \mathcal{D} \) is reenactment module, which will be introduced in the following section. Finally, by concatenate \( \mathcal{G}_{static} \) and \( \mathcal{G}_{dynamic} \), we got the completed 3D Gaussian \( \mathcal{G} = [\mathcal{G}_{static} , \mathcal{G}_{dynamic}]\) of avatar. 

\subsection{Reenactment Module}
In this subsection we give more details for reenactment module. The reenactment module first employs several layers of MLP to upscale the input control parameter vector \(\alpha_d\), and then reshapes it into a 2D feature map. Next, it performs further feature extraction via 2D convolution and predicts the positions of the dynamic 3D Gaussian. During inference, following the one-shot reconstruction from source image \({I_s}\), all subsequent expression reenactments are processed solely through the reenactment module. Due to its extremely lightweight design, the reenactment module enables highly efficient synthesis and rendering. 

\subsection{Texture Restoration Module}

Most 3D methods estimate pose and scale from monocular videos/images, causing camera pose errors and high-frequency texture artifacts. In contrast, 2D methods use a warping field for a pseudo-3D transformation that visually approximates true 3D motion, their simpler motion patterns are easier to learn. To address 3D artifacts, we employed a lightweight texture restoration module after the 3D Gaussian Splatting stage, using a StyleGAN2 Generator with SFT modulation. Following the Live Portrait \cite{guo2024liveportrait} and GFPGAN \cite{wang2021towards} approachs, we synthesized about 70K training samples using FFHQ as facial appearance and VFHQ as the driving video.


\subsection{Training Strategy and Loss Functions}
Our training process is divided into two stages: global pretraining and fine-tuning for texture restoration module. Global pretraining employs VFHQ as the training dataset. For each training sample, two video frames are randomly selected from the same video sequence to serve respectively as the source image and the drive image. The corresponding driving frame control parameters \(\alpha_d\) and the source image \(I_s\) are fed into the reconstruction module  to predict the sync image \(\hat{I_d}\). Since the source image and drive image belong to the same person, the drive image can be used as the target image to provide pixel-level supervision for the sync image \(\hat{I_d}\). The loss function for global training mainly consists of a perceptual loss \(L_{lpips}\) and an loss \(L_1 = || \hat{I_d} - I_d ||_1 \). 

\begin{equation} 
L_{pretraing} = \lambda_{1} L_{lpips} + \lambda_{2} L_1,
\end{equation} 
Where \(\lambda_{1} = 0.01 \) and \(\lambda_{2} = 1 \).

During texture restoration module fine-tuning, we freeze all parameters except those of the texture restoration module and train on 70K synthetic data samples. The fine-tuning loss is also composed of a perceptual loss and an \(L_1 \) loss, with additional supervision loss functions \( L_{eye\_teeth} \) applied to specific regions such as the eyes and teeth. Specifically, the \( L_{eye\_teeth} \) are implemented as follows. First, the eye and mouth bounding box regions in \( I_d\) are estimated using the landmarks calculated by the facial landmarks detector. Then, the perceptual loss for the eye and mouth regions is calculated separately and summed.

\begin{equation} 
L_{finetune} = \lambda_{1} L_{lpips} + \lambda_{2} L_1  + \lambda_{3} L_{eye\_teeth}
\end{equation} 
Where \(\lambda_{1}\) and \(\lambda_{2} \) keep consistent values as before, and \( \lambda_{3} = 0.05\).
\section{Experiment}
\subsection{Data Preprocessing.}
For both training and inference data, we adopt a consistent preprocessing procedure. First, the background is removed and the foreground subject is preserved by applying a matting algorithm. Subsequently, the 3D landmarks are estimated and utilized alongside a predefined standard model to compute the head pose. Thereafter, the camera pose relative to the standard model is derived from the computed head pose. 


\subsection{Experimental Settings.}
In global training, VFHQ is used as the training set. For each video frame sequence, one frame is sampled at an interval of five frames, forming a sampled subset as the training data. In total, over 600,000 frames were extracted from more than 15,000 videos. During fine-tuning of the restoration module, a subset of 7K person identity data is first filtered from FFHQ. Then, for each identity, 10 videos are randomly selected from VFHQ as driving videos. Finally, based on Live Portrait \cite{guo2024liveportrait} and GFPGAN \cite{wang2021towards},  70K video clips were generated, and one frame was sampled every five frames for a clip, yielding a final total of 1.5 million frames. All training data were resized to 512×512 resolution. The test data consist of 50 videos from the original test split in VFHQ, as well as all 365 videos from the HDTF dataset.

\begin{table*}[htbp]
\vspace{-0.8cm}
\small
\setlength{\tabcolsep}{3pt}
\renewcommand{\arraystretch}{1.1}
\centering

\hfill
\begin{minipage}{1\linewidth}
\begin{tabular}{l|c|cccccccc|ccc}

\toprule
\hline
\multirow{1}{*}{\textbf{Backbone}} & 
\multirow{1}{*}{\textbf{Params}} & 
\multicolumn{8}{c}{\textbf{Self Reenactment}} & 
\multicolumn{3}{|c}{\textbf{Cross Reenactment}} \\
& &
\rotatebox[origin=c]{0}{\textbf{CSIM↑}} & 
\rotatebox[origin=c]{0}{\textbf{LPIPS↓}} & 
\rotatebox[origin=c]{0}{\textbf{SSIM↑}} & 
\rotatebox[origin=c]{0}{\textbf{PSNR↑}} & 
\rotatebox[origin=c]{0}{\textbf{FLMD↓}} & 
\rotatebox[origin=c]{0}{\textbf{MLMD↓}} & 
\rotatebox[origin=c]{0}{\textbf{APC↑}} & 
\rotatebox[origin=c]{0}{\textbf{AEC↑}} & 
\rotatebox[origin=c]{0}{\textbf{CSIM↑}} & 
\rotatebox[origin=c]{0}{\textbf{APC↑}} &
\rotatebox[origin=c]{0}{\textbf{AEC↑}} \\
\hline
\textbf{Dinov2-vitb14 }  & 0.1B 
&0.879	&0.151	&0.771	&20.826	&0.833	&1.209	&0.981	&0.953	&0.456	&0.810	&0.388 \\
\textbf{Dinov2-vitg14 }  & 1.1B
&\underline{0.881}	&\underline{0.151}	&\underline{0.774}	&\underline{20.932}	&\underline{0.829}	&\underline{1.173}	&\underline{0.982}	&\underline{0.959}	&\underline{0.456}	&\underline{0.810}	&\underline{0.397} \\
\textbf{Webssl-dino7b-full8b-518 }  & 7.0B
&\textbf{0.884}	&\textbf{0.146}	&\textbf{0.785}	&\textbf{21.084}	&\textbf{0.811}	&\textbf{1.172}	&\textbf{0.987}	&\textbf{0.964}	&\textbf{0.466}	&\textbf{0.817}	&\textbf{0.407}\\
\bottomrule
\end{tabular}
\end{minipage}
\vspace{0.1cm}
\caption{Quantitative comparison of different backbone on HDTF. We highlight the \textbf{best} and \underline{second best} results.} 
\label{tab:comparsion_backbone}
\end{table*}

\begin{table*}[t]
\centering
\setlength{\tabcolsep}{4pt}
\begin{tabular}{l|cccccccc}
\toprule
\hline
Method &StyleHeat &Real3D   &P4D   &P4D-v2  &LivePortrait &GAGavatar  &LAM & RAR \\
\hline 
FPS & 19.82 &4.55   &9.51   &9.62  &77 &67.12  &\textbf{280.96} & \underline{90} \\
\hline
\end{tabular}

\caption{Running time of driving and rendering measured in FPS and tested on A100 GPU with \(512 \times 512\) resolution outputs. All results exclude the time for reconstruction
 and driving parameters estimation which can be calculated in advance. We highlight the \textbf{best} and \underline{second best} results.}
 \label{tab:Inference Speed}
\vspace{-0.2cm}
\end{table*}

\begin{figure}[t]
\vspace{-0.8cm}
\centering
\includegraphics[width=1\columnwidth]{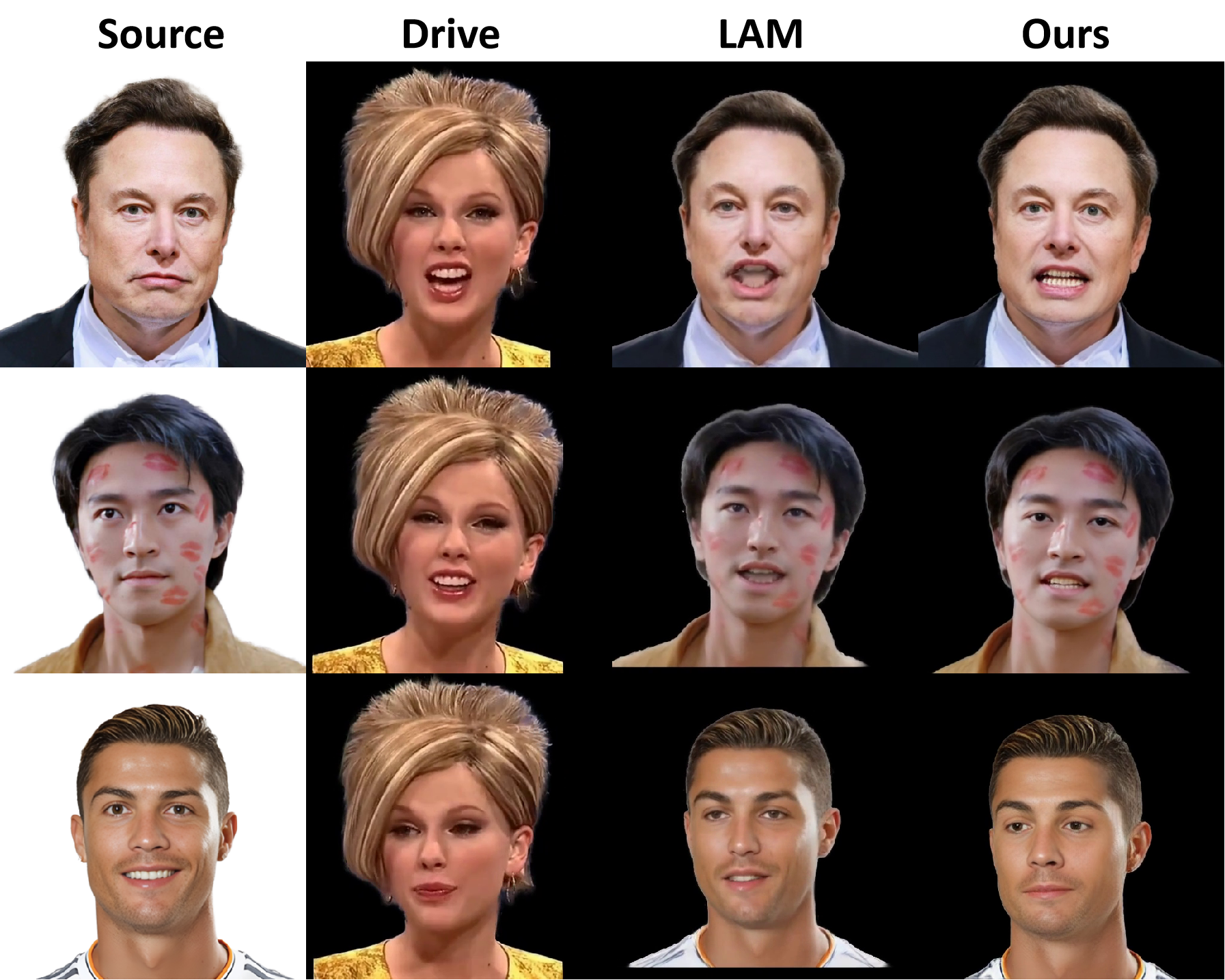} 
\caption{Visual comparison between LAM\cite{he2025lam} and RAR.}
\label{fig:comparison_of_LAM_and_Ours}
\vspace{-0.3cm}
\end{figure}

\begin{figure}[t]
\vspace{-0.8cm}
\centering
\includegraphics[width=0.85\columnwidth]{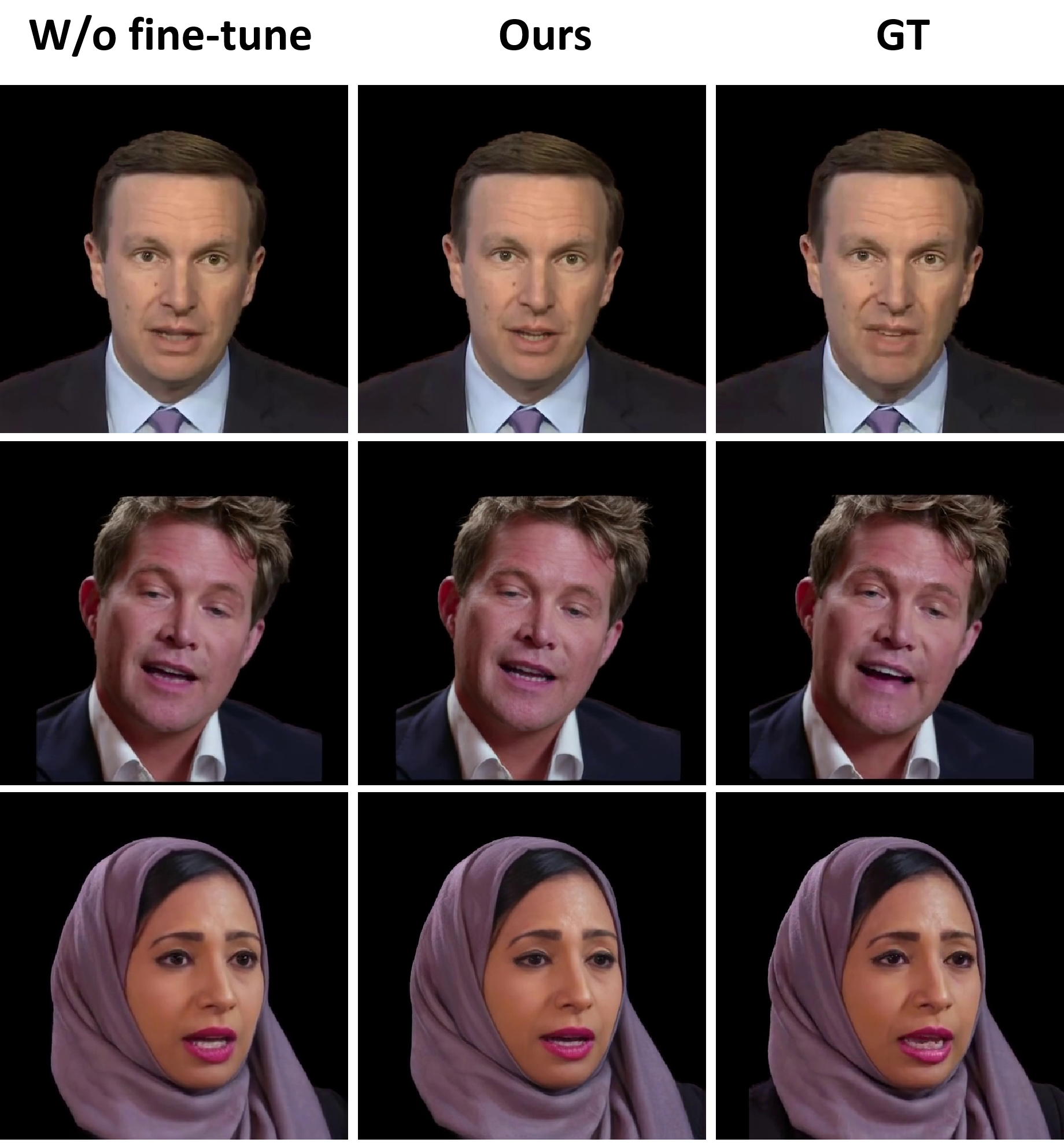} 
\caption{With texture restoration module and synthetic data, more detailed textures in the tooth region can be reconstructed}
\label{fig:texture_restoration}
\vspace{-0.3cm}
\end{figure}

\subsection{Implementation Details.}
We implemented the entire experiment using PyTorch with 8 H200 GPUs. In global training we set a global batch size of 128 and ran training for 500K iterations. The learning rate was 1e-4, and we used the ADAM optimizer. Next, we fine-tuned the restoration module. We kept the batch size and optimizer unchanged. We set the learning rate to 5e-5. Regarding the weighting coefficients of the loss functions, the perceptual loss is weighted at 0.01 and the L1 loss at 1. Additionally, the loss function for specific regions is weighted at 0.05.

\subsection{Evaluation Metrics.}
In the evaluation process, we focus on performance metrics for self-identity reenactment and cross-identity reenactment. For self-reenactment, the drive images serve as the ground truth, establishing a supervised evaluation. We assess the quality of synthesis along three dimensions: image quality, identity similarity, and expression/pose similarity. Regarding image quality, we adopt three quantitative metrics: Peak Signal-to-Noise Ratio (PSNR), Structural Similarity Index (SSIM), and Learned Perceptual Image Patch Similarity (LPIPS). These metrics enable an effective comparison between synthesized and drive images. To evaluate identity similarity (CSIM), we calculate the cosine distance of face recognition features based on the methodology of \cite{deng2024portrait4d}. For assessing the realism of expressions and poses, we use the average expression cosine distance (AEC) and average pose cosine distance (APC), as determined by a 3D Morphable Model (3DMM) estimator. Additionally, we employ the facial keypoint detector introduced to measure the facial landmark mean distance (FLMD) and the mouth landmark mean distance (MLMD), which offer deeper insights into the accuracy of the driving controls in our animation. In the case of cross-identity reenactment, ground truth are not available. Therefore, we rely solely on CSIM, AEC, and APC for evaluation. Except for minor changes in the AEC and APC calculation process, these metrics are consistent with those used in previous studies \cite{deng2024portrait4d,deng2024portrait4dv2,chu2024generalizable,chu2024gpaavatar,he2025lam}.

\section{Results}
\subsection{Quantitative Results.}
We evaluated RAR on the VFHQ and HDTF datasets. The results on these datasets are presented in Tab.\ref{tab:comparsion_VFHQ} and Tab.\ref{tab:comparsion_HDTF}, respectively. As shown in the tables, our method achieved the best image reconstruction quality, as evidenced by the PSNR, SSIM, and LPIPS metrics. Additionally, we maintained strong identity consistency, which is reflected in the CSIM metric. Moreover, we obtained accurate expressions and poses consistent with the driving image, as indicated by the FLMD, MLMD, AEC and APC metrics.

\subsection{Qualitative Results.}
Fig.\ref{fig:cross reenactment} compares the cross reenactment results of RAR with other approaches on both the VFHQ and HDTF datasets. Compared to previous methods, our approach achieves superior reconstruction details in texture, better preserves identity consistency, and expressions and poses more aligned with the driving image. It is also evident that after fine-tuning with texture repair, our model shows significant improvements in high-frequency texture regions (such as teeth and eyes).

\subsection{Inference Speed.}

During the inference phase, latency is primarily composed of two parts: drive and render. As shown in Fig.\ref{fig:full_pipe}, the modules involved in the drive and render stages are indicated by dashed boxes. The drive portion consists solely of the reenactment module, which involves only a few MLP and Conv2D operations, resulting in a negligibly short running time of less than 1 ms. The main inference time of RAR is concentrated in the render stage. Benefiting from the learnability of synthetic data, and with reference to StyleGAN2 \cite{karras2020stylegan}, we stacked three SFT modulations to implement a lightweight texture completion module. This configuration allows the entire render stage to run in approximately 9 ms, while ensuring high-quality reconstruction of high-frequency textures. Ultimately, the combined inference speed of the drive and render stages reaches 90 FPS. In Tab.\ref{tab:Inference Speed}, we compare RAR with several previous algorithms. The inference speed of RAR is second only to LAM, however our method can achieve significantly superior visual results as shown in Fig.\ref{fig:comparison_of_LAM_and_Ours}. Additionally since LAM did not provide a complete custom testing sample creation process, we could only align with the test cases provided by LAM for comparison. 



\


\begin{table}[htbp]
\vspace{-0.3cm}
\small
\setlength{\tabcolsep}{6pt}
\renewcommand{\arraystretch}{1.1}
\centering

\hfill
\begin{minipage}{1\linewidth}
\begin{tabular}{l|ccc}

\toprule
\hline
\multirow{1}{*}{\textbf{Control Priors}} &

\rotatebox[origin=c]{0}{\textbf{CSIM↑}} & 
\rotatebox[origin=c]{0}{\textbf{APC↑}} &
\rotatebox[origin=c]{0}{\textbf{AEC↑}} \\
\hline
\textbf{FLAME \cite{li2017learning}}  
&\underline{0.451}	&\underline{0.801}	&\underline{0.399} \\
\textbf{PDFGC \cite{wang2023progressive}}  
&\textbf{0.466}	&\textbf{0.817}	&\textbf{0.407} \\

\bottomrule
\end{tabular}
\end{minipage}
\caption{Quantitative comparison of different control priors with cross-identity reenactment setting on HDTF. We highlight the \textbf{best} and \underline{second best} results.} 
\vspace{-0.5cm}
\label{tab:comparsion_control_priors}
\end{table}

\section{Ablation Studies}
\subsection{Pre-trained Backbone.}
Based on the HDTF test sets, we compared the impact of different-scale models used as backbones in our method. As shown in Tab.\ref{tab:comparsion_backbone}, as the scale of the pre-trained backbone increases, the performance of our algorithm correspondingly improves. Although our current backbone is based on Webssl-dino7b-full8b-518 \cite{fan2025webssl}, the architecture which decouples reconstruction and reenactment can theoretically support larger-scale pre-trained models as backbones, conforming to the performance enhancement indicated by the scaling law \cite{kaplan2020scaling} while still maintaining the current high driving efficiency. 

\subsection{Texture Restoration.}
We compared the visual results to evaluate the effect differences caused by incorporating a texture restoration module and training with synthetic data. As shown in Fig.\ref{fig:texture_restoration}, after fine-tuning, more detailed textures in the tooth region can be reconstructed. This confirms the effectiveness of texture restoration.

\subsection{Comparison of Control Priors}
We conducted a comparative experiment to show that using PDFGC \cite{wang2023progressive} as the control prior is better than using FLAME. We kept the model architecture and training method the same. We trained two models: one with PDFGC as the control prior and one with FLAME. We then evaluated the cross reenactment performance on the HDTF datasets. Table \ref{tab:comparsion_control_priors} shows that the PDFGC-based approach performs better than the FLAME-based method according to the evaluation metrics.

\bibliography{aaai2026}

\begin{thebibliography}{51}
\providecommand{\natexlab}[1]{#1}

\bibitem[{Baevski and others.(2020)}]{baevski2020wav2vec}
Baevski, A.; and others. 2020.
\newblock wav2vec 2.0: A framework for self-supervised learning of speech representations.
\newblock In \emph{NeurIPS}.

\bibitem[{Bai and others.(2023)}]{bai2023high}
Bai, Y.; and others. 2023.
\newblock High-fidelity Facial Avatar Reconstruction from Monocular Video with Generative Priors.
\newblock \emph{in CVPR}.

\bibitem[{Blanz and others.(1999)}]{blanz1999morphable}
Blanz, V.; and others. 1999.
\newblock A Morphable Model for the Synthesis of 3D Faces.
\newblock \emph{in SIGGRAPH}.

\bibitem[{Burkov and others.(2020)}]{burkov2020neural}
Burkov, E.; and others. 2020.
\newblock Neural Head Reenactment with Latent Pose Descriptors.
\newblock \emph{in CVPR}.

\bibitem[{Chen and others.(2025)}]{chen2025hunyuanvideo}
Chen, Y.; and others. 2025.
\newblock HunyuanVideo-Avatar: High-Fidelity Audio-Driven Human Animation for Multiple Characters.
\newblock \emph{in ArXiv}.

\bibitem[{Chu and others.(2024{\natexlab{a}})}]{chu2024generalizable}
Chu, X.; and others. 2024{\natexlab{a}}.
\newblock Generalizable and Animatable Gaussian Head Avatar.
\newblock \emph{in NeurIPS}.

\bibitem[{Chu and others.(2024{\natexlab{b}})}]{chu2024gpaavatar}
Chu, X.; and others. 2024{\natexlab{b}}.
\newblock GPAvatar: Generalizable and Precise Head Avatar from Image(s).
\newblock \emph{in ICLR}.

\bibitem[{Cui and others.(2024)}]{cui2024hallo2}
Cui, J.; and others. 2024.
\newblock Hallo2: Long-Duration and High-Resolution Audio-Driven Portrait Image Animation.
\newblock \emph{in CoRR}.

\bibitem[{Deng and others.(2024{\natexlab{a}})}]{deng2024portrait4d}
Deng, Y.; and others. 2024{\natexlab{a}}.
\newblock Portrait4D: Learning One-Shot 4D Head Avatar Synthesis using Synthetic Data.
\newblock \emph{in CVPR}.

\bibitem[{Deng and others.(2024{\natexlab{b}})}]{deng2024portrait4dv2}
Deng, Y.; and others. 2024{\natexlab{b}}.
\newblock Portrait4D-v2: Pseudo MultiView Data Creates Better 4D Head Synthesizer.
\newblock \emph{in ArXiv}.

\bibitem[{Fan and others.(2025)}]{fan2025webssl}
Fan, D.; and others. 2025.
\newblock Scaling language-free visual representation learning.
\newblock \emph{in ArXiv}.

\bibitem[{Gafni and others.(2021)}]{gafni2021dynamic}
Gafni, G.; and others. 2021.
\newblock Dynamic neural radiance fields for monocular 4d facial avatar reconstruction.
\newblock \emph{in CVPR}.

\bibitem[{Gerig and others.(2018)}]{gerig2018morphable}
Gerig, T.; and others. 2018.
\newblock Morphable Face Models - An Open Framework.
\newblock \emph{in FG}.

\bibitem[{Goodfellow and others.(2014)}]{goodfellow2014generative}
Goodfellow, I.~J.; and others. 2014.
\newblock Generative Adversarial Nets.
\newblock \emph{in NeurIPS}.

\bibitem[{Guo and others.(2024)}]{guo2024liveportrait}
Guo, J.; and others. 2024.
\newblock LivePortrait: Efficient Portrait Animation with Stitching and Retargeting Control.
\newblock \emph{in CoRR}.

\bibitem[{Han and others.(2024)}]{han2024cvthead}
Han, H.; and others. 2024.
\newblock CVTHead: One-shot Controllable Head Avatar with Vertex-feature Transformer.
\newblock \emph{in WACV}.

\bibitem[{He and others.(2025)}]{he2025lam}
He, Y.; and others. 2025.
\newblock LAM: Large Avatar Model for One-shot Animatable Gaussian Head.
\newblock \emph{in ArXiv}.

\bibitem[{Hong and others.(2022)}]{hong2022depth}
Hong, F.-T.; and others. 2022.
\newblock Depth-Aware Generative Adversarial Network for Talking Head Video Generation.
\newblock \emph{in CVPR}.

\bibitem[{Isola and others.(2017)}]{isola2017image}
Isola, P.; and others. 2017.
\newblock Image-to-Image Translation with Conditional Adversarial Networks.
\newblock \emph{in CVPR}.

\bibitem[{Kaplan and others.(2020)}]{kaplan2020scaling}
Kaplan, J.; and others. 2020.
\newblock Scaling laws for neural language models.
\newblock \emph{in ArXiv}.

\bibitem[{Karras and others.(2019)}]{karras2019style}
Karras, T.; and others. 2019.
\newblock A Style-Based Generator Architecture for Generative Adversarial Networks.
\newblock \emph{in CVPR}.

\bibitem[{Karras and others.(2020)}]{karras2020stylegan}
Karras, T.; and others. 2020.
\newblock Analyzing and Improving the Image Quality of StyleGAN.
\newblock \emph{in CVPR}.

\bibitem[{Kerbl and others.(2023)}]{kerbl2023gaussian}
Kerbl, B.; and others. 2023.
\newblock 3D Gaussian Splatting for Real-Time Radiance Field Rendering.
\newblock \emph{in ACM TOG}.

\bibitem[{Khakhulin and others.(2022)}]{khakhulin2022realistic}
Khakhulin, T.; and others. 2022.
\newblock Realistic One-Shot Mesh-Based Head Avatars.
\newblock \emph{in ECCV}.

\bibitem[{Li and others.(2017)}]{li2017learning}
Li, T.; and others. 2017.
\newblock Learning a model of facial shape and expression from 4D scans.
\newblock \emph{in ACM TOG}.

\bibitem[{Li and others.(2023)}]{li2023generalizable}
Li, X.; and others. 2023.
\newblock Generalizable One-shot 3D Neural Head Avatar.
\newblock \emph{in NeurIPS}.

\bibitem[{Mildenhall and others.(2020)}]{mildenhall2020nerf}
Mildenhall, B.; and others. 2020.
\newblock NeRF: Representing Scenes as Neural Radiance Fields for View Synthesis.
\newblock \emph{in ECCV}.

\bibitem[{Oquab and others.(2024)}]{oquab2024dinov2}
Oquab, M.; and others. 2024.
\newblock DINOv2: Learning Robust Visual Features without Supervision.
\newblock \emph{in TMLR}.

\bibitem[{Paysan and others.(2009)}]{paysan20093dface}
Paysan, P.; and others. 2009.
\newblock A 3D Face Model for Pose and Illumination Invariant Face Recognition.
\newblock \emph{in AVSS}.

\bibitem[{Prajwal and others.(2020)}]{prajwal2020lip}
Prajwal, K.; and others. 2020.
\newblock A lip sync expert is all you need for speech to lip generation in the wild.
\newblock In \emph{ACM MM}.

\bibitem[{Qian and others.(2024)}]{qian2024gaussianavatars}
Qian, S.; and others. 2024.
\newblock GaussianAvatars: Photorealistic Head Avatars with Rigged 3D Gaussians.
\newblock \emph{in CVPR}.

\bibitem[{Sun and others.(2023)}]{sun2023next3d}
Sun, J.; and others. 2023.
\newblock Next3D: Generative Neural Texture Rasterization for 3D-Aware Head Avatars.
\newblock \emph{in CVPR}.

\bibitem[{Sun and others.(2024)}]{sun2024cgof}
Sun, K.; and others. 2024.
\newblock CGOF++: Controllable 3D Face Synthesis With Conditional Generative Occupancy Fields.
\newblock \emph{in IEEE Trans. Pattern Anal. Mach. Intell.}

\bibitem[{Tang and others.(2024)}]{tang20243dfaceshop}
Tang, J.; and others. 2024.
\newblock 3DFaceShop: Explicitly Controllable 3D-Aware Portrait Generation.
\newblock \emph{in TVCG}.

\bibitem[{Tian and others.(2024)}]{tian2024emo}
Tian, L.; and others. 2024.
\newblock EMO: Emote Portrait Alive Generating Expressive Portrait Videos with Audio2Video Diffusion Model Under Weak Conditions.
\newblock \emph{in ECCV}.

\bibitem[{Wang and others.(2023)}]{wang2023progressive}
Wang, D.; and others. 2023.
\newblock Progressive Disentangled Representation Learning for Fine-Grained Controllable Talking Head Synthesis.
\newblock \emph{in CVPR}.

\bibitem[{Wang and others.(2021{\natexlab{a}})}]{wang2021oneshot}
Wang, T.-C.; and others. 2021{\natexlab{a}}.
\newblock One-Shot Free-View Neural Talking-Head Synthesis for Video Conferencing.
\newblock \emph{in CVPR}.

\bibitem[{Wang and others.(2021{\natexlab{b}})}]{wang2021towards}
Wang, X.; and others. 2021{\natexlab{b}}.
\newblock Towards real-world blind face restoration with generative facial prior.
\newblock \emph{in CVPR}, 9168--9178.

\bibitem[{Xu and others.(2023)}]{xu2023pv3d}
Xu, E.~Z.; and others. 2023.
\newblock PV3D: A 3D Generative Model for Portrait Video Generation.
\newblock \emph{in ICLR}.

\bibitem[{Xu and others.(2020)}]{xu2020deep3d}
Xu, S.; and others. 2020.
\newblock Deep 3D Portrait From a Single Image.
\newblock \emph{in CVPR}.

\bibitem[{Xu and others.(2024)}]{xu2024gaussianhead}
Xu, Y.; and others. 2024.
\newblock Gaussian Head Avatar: Ultra High-Fidelity Head Avatar via Dynamic Gaussians.
\newblock \emph{in CVPR}.

\bibitem[{Ye and others.(2024)}]{ye2024real3d}
Ye, Z.; and others. 2024.
\newblock Real3d-portrait: One-shot realistic 3d talking portrait synthesis.
\newblock \emph{in ArXiv}.

\bibitem[{Ye and others.(2025)}]{ye2025realistic}
Ye, Z.; and others. 2025.
\newblock Realistic Real-Time Talking Head Synthesis with Grid Encoding and Progressive Conditioning.
\newblock \emph{in ICASSP}.

\bibitem[{Yin and others.(2022)}]{yin2022styleheat}
Yin, F.; and others. 2022.
\newblock StyleHEAT: One-Shot High-Resolution Editable Talking Face Generation via Pre-trained StyleGAN.
\newblock \emph{in ECCV}.

\bibitem[{Yu and others.(2023)}]{yu2023nofa}
Yu, W.; and others. 2023.
\newblock NOFA: NeRF-based One-shot Facial Avatar Reconstruction.
\newblock \emph{in SIGGRAPH}.

\bibitem[{Zakharov and others.(2019)}]{zakharov2019fewshot}
Zakharov, E.; and others. 2019.
\newblock Few-Shot Adversarial Learning of Realistic Neural Talking Head Models.
\newblock \emph{in ICCV}.

\bibitem[{Zhang and others.(2023)}]{zhang2023metaportrait}
Zhang, B.; and others. 2023.
\newblock MetaPortrait: Identity-Preserving Talking Head Generation with Fast Personalized Adaptation.
\newblock \emph{in CVPR}.

\bibitem[{Zhang and others.(2024)}]{zhang2024musetalk}
Zhang, Y.; and others. 2024.
\newblock Musetalk: Real-time high quality lip synchronization with latent space inpainting.
\newblock \emph{in ArXiv}.

\bibitem[{Zheng and others.(2023)}]{zheng2023pointavatar}
Zheng, Y.; and others. 2023.
\newblock PointAvatar: Deformable Point-Based Head Avatars from Videos.
\newblock \emph{in CVPR}.

\bibitem[{Zhuang and others.(2022)}]{zhuang2022controllable}
Zhuang, P.; and others. 2022.
\newblock Controllable Radiance Fields for Dynamic Face Synthesis.
\newblock \emph{in 3DV}.

\bibitem[{Zielonka and others.(2023)}]{zielonka2023instant}
Zielonka, W.; and others. 2023.
\newblock Instant Volumetric Head Avatars.
\newblock \emph{in CVPR}.

\end{thebibliography}

\section{Supplementary}
\subsection{A. Audio Driven}
Our proposed method, RAR, enables independent control of facial features such as the eyes and mouth in a single-frame portrait. To further validate the excellent control capability and expand the application scope of our approach, we designed an audio2motion (audio to motion) model that maps the input audio to the PDFGC mouth parameters. The mouth parameters generated by audio2motion control the mouth shape of a 3D Gaussian head, ultimately realizing speech-driven lip synchronization of the 3D Gaussian head.

Specifically, the audio2motion model is implemented as a modular neural network that maps an audio sequence to a motion sequence. The architecture mainly consists of three components. First, a pre-trained Wav2Vec2.0 \cite{baevski2020wav2vec} model is employed as the audio encoder, with its feature extraction layers frozen. The contextual output is then projected to a configurable hidden dimension via a linear layer. Second, the intermediate audio features are further processed by an audio encoder module, which optionally incorporates identity or category information to facilitate personalized modeling. Third, in the decoding stage, the architecture utilizes multiple sequentially arranged ConvNormRelu blocks. The block comprises a 1D convolution, layer normalization, and ReLU activation, with a final 1D convolution mapping the hidden state to the output dimension.

During training, we collected a dataset comprising 1,000 professional single-speaker lecture videos. From each frame, mouth features were extracted through PDFGC, and the corresponding audio segments served as inputs to constitute paired training samples. This dataset was subsequently used to train the audio2motion model which employed a mean square error (MSE) loss to constrain the predicts to the ground-truth.

\begin{table}[htbp]
\small
\setlength{\tabcolsep}{13pt}
\renewcommand{\arraystretch}{1.1}
\centering

\hfill
\begin{minipage}{1\linewidth}
\begin{tabular}{l|cc}

\toprule
\hline
\textbf{Method} & 
\rotatebox[origin=c]{0}{\textbf{Sync Score↑}} & 
\rotatebox[origin=c]{0}{\textbf{CSIM↑}} \\
\hline

\textbf{MuseTalk}  
&6.33	&0.302 \\
\textbf{Wav2Lip}  
&\textbf{7.58}	&0.306	 \\

\textbf{HunyuanVideoAvatar}  
&\underline{7.34}	&\underline{0.336}	 \\

\textbf{RAR}  
&6.52	&\textbf{0.351}	 \\

\bottomrule
\end{tabular}
\end{minipage}
\caption{\textbf{Quantitative Results of Audio-driven.} We highlight the \textbf{best} and \underline{second best} results.} 
\label{tab:audio_driven}
\end{table}

Furthermore, quantitative experiments verified that our approach, when extended to audio-driven tasks, achieves highly competitive results. We randomly selected 50 task IDs and 50 audio clips from HDTF for cross-driving experiments. The accuracy of the lip synchronization was measured using the sync score \cite{prajwal2020lip}, while identity consistency was evaluated using the CSIM metric. For comparison, we selected Wav2Lip, MuseTalk, and HunyuanVideoAvatar as baseline methods. As shown in Table \ref{tab:audio_driven}, detailed comparative results are provided: Wav2Lip \cite{prajwal2020lip}, which directly uses the sync score as a supervisory loss, achieved the highest lip accuracy score. HunyuanVideoAvatar \cite{chen2025hunyuanvideo}, benefiting from large-scale parameters and data, attained a sync score second only to that of Wav2Lip. And our method achieved a sync score slightly better than musetalk \cite{zhang2024musetalk}, though lower than both Wav2Lip and HunyuanVideoAvatar. On the other hand, unlike other methods, our approach, benefiting from enhanced 3D priors, performed best in terms of the CSIM metric. Therefore, RAR can be directly extended to audio-driven tasks without additional training while achieving highly competitive results, further validating its strong independent control over facial features in a single-frame portrait.

\subsection{B. Comparison Methods}
We carefully selected state-of-the-art approaches for comparison, including 2D end-to-end methods such as StyleHeat \cite{yin2022styleheat} and LivePortrait \cite{guo2024liveportrait}, as well as methods based on 3D prior structures such as CVTHead \cite{han2024cvthead}, GPAvatar \cite{chu2024gpaavatar}, Real3DPortrait \cite{ye2024real3d}, Portrait4D \cite{deng2024portrait4d}, Portrait4D-v2 \cite{deng2024portrait4dv2}, and GAGAvatar \cite{chu2024generalizable}. For each method, we reproduced the results using the official open-source projects and models.

\subsection{C. Structure Details}
In this part, we introduce more details of RAR.

\begin{figure*}[t]
    \includegraphics[width=1\linewidth]{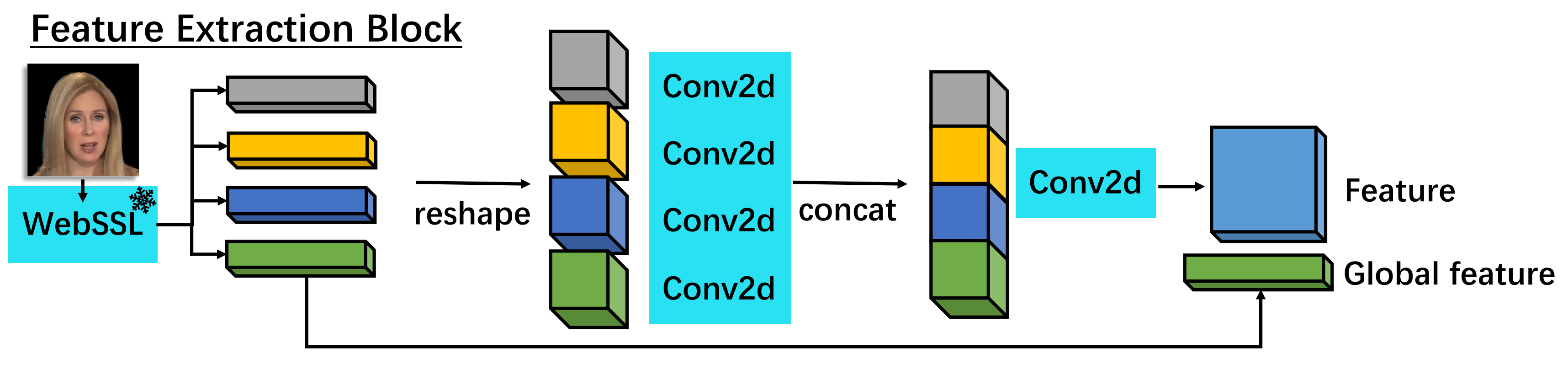}
    \caption{The module uses a pre-trained WebSSL model to extract multi-granularity embeddings, reshaping them into 2D maps. After Conv2D processing, these maps are concatenated and fused by a final Conv2D layer, while the last WebSSL layer provides the global feature.}
    \label{fig:feature_extraction_block}
\end{figure*}
\textbf{Feature Extraction Module.} The feature extraction module first leverages a pre-trained WebSSL model to extract embeddings at varying granularities. These embeddings are subsequently reshaped into 2D feature maps. After processing through convolutional layers (Conv2D), the multi-granularity feature maps are concatenated along the channel dimension. A final Conv2d layer then integrates these concatenated features, yielding a fused feature map. Additionally, the embedding from the final layer of the WebSSL model is selected as the global feature. The more intuitive process is illustrated in the Fig.\ref{fig:feature_extraction_block}.

\begin{figure*}[t]
    \includegraphics[width=1\linewidth]{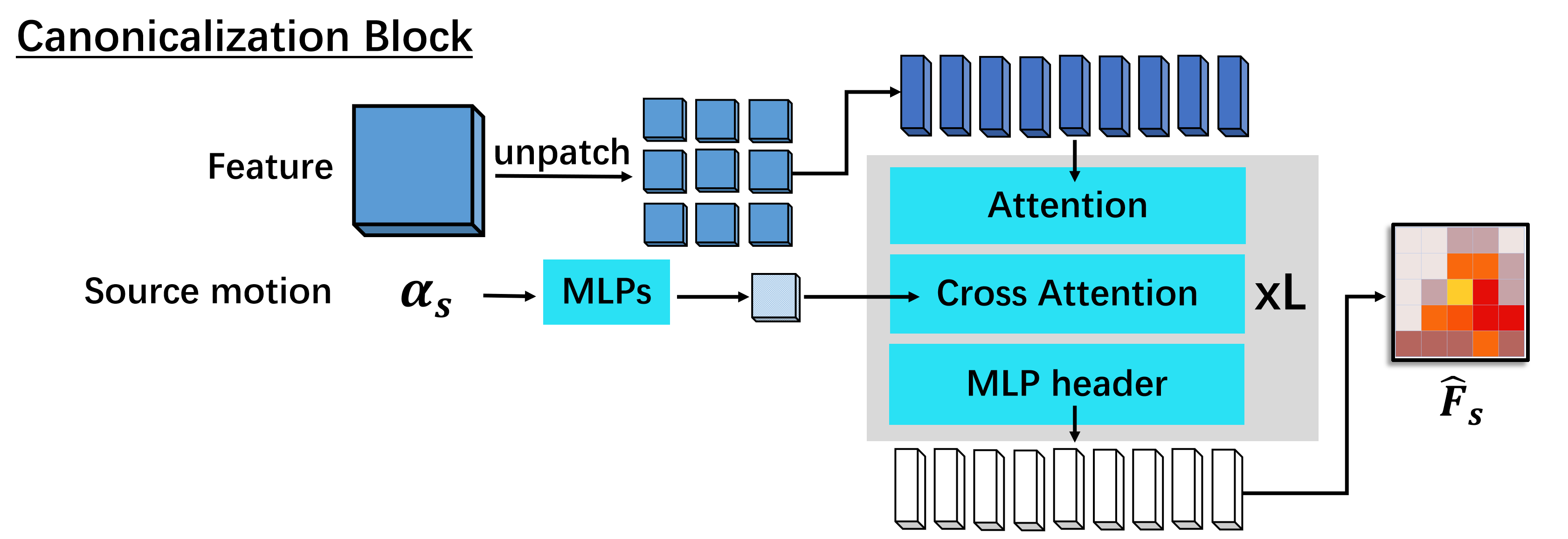}
    \caption{Canonicalization Block neutralizes the expression of features on the source image via source motion. The fused features are split into patches, reshaped into 1D token sequences, and processed through \(L\) transformer blocks. Finally, tokens are rearranged to reconstruct the canonical feature map, with the source motion vector transformed by a multi-layer MLP and injected as conditioning for Cross-Attention.}
    \label{fig:canonicalization_block}
\end{figure*}

\textbf{Canonicalization Block.} The Canonicalization Block achieves expression neutralized feature on the source image by leveraging the source motion. The fused feature is initially partitioned into patches. Each patch is reshaped into a 1D token sequence. These tokens are processed through \(L\) stacked transformer blocks, each comprising Multi-Head Attention, Cross-Attention, and a Multilayer Perceptron (MLP) head layer. Finally, the processed tokens are rearranged (unpatched) to reconstruct the canonicalized feature map. Crucially, the source motion vector undergoes transformation via a multi-layer MLP projection and is injected as the conditioning input for the Cross-Attention within this process. For more intuitive process can be found in the Fig.\ref{fig:canonicalization_block}.

\begin{figure*}[t]
    \vspace{-0.8cm}
    \includegraphics[width=1\linewidth]{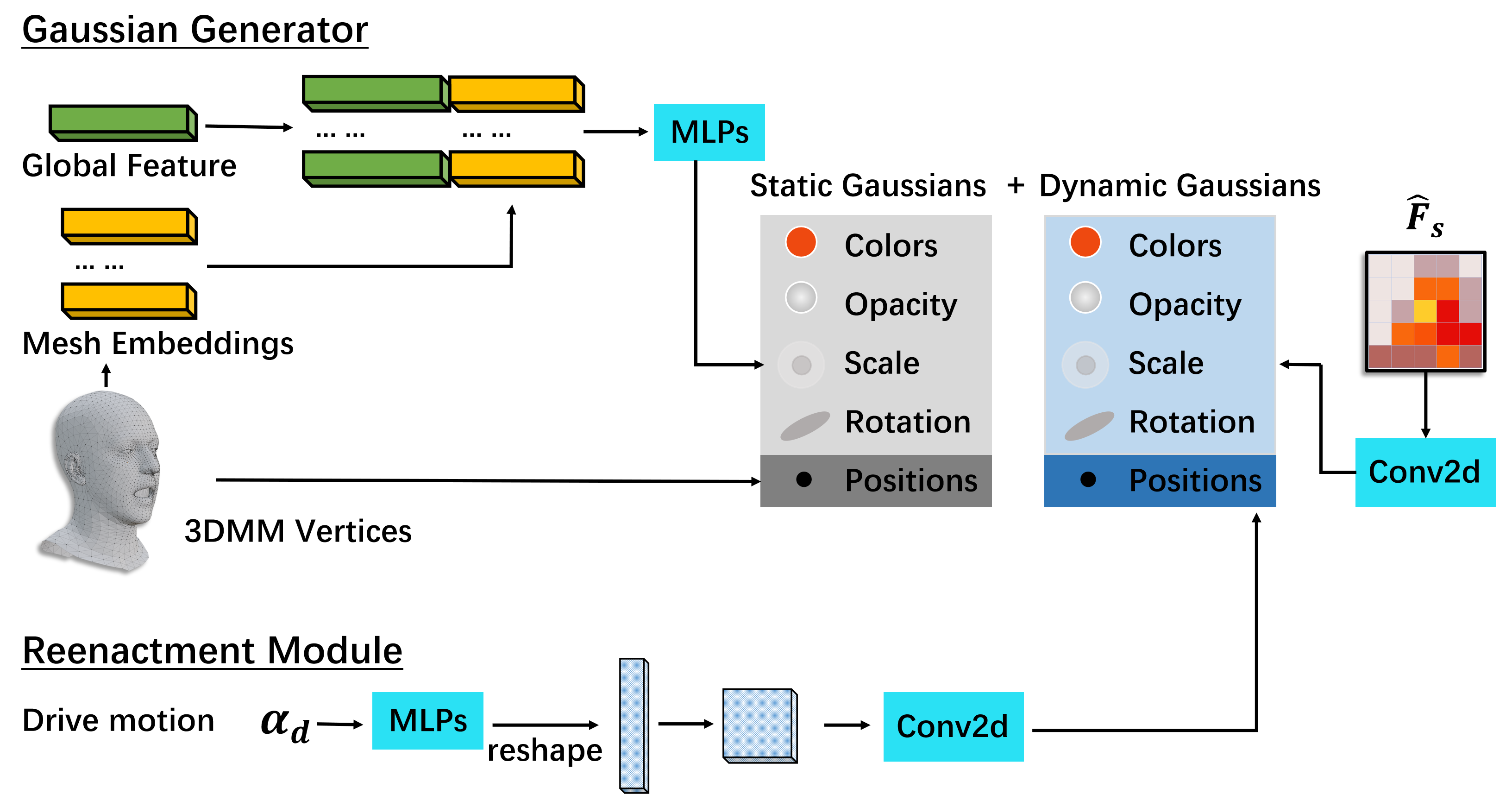}
    \caption{The Gaussian Generator produces two parts: the static and dynamic Gaussians. The static Gaussian uses 3DMM vertices for positions, while its appearance is predicted by concatenating a global feature with a mesh embedding of those vertices and processing them through MLP layers. The dynamic Gaussian's appearance is predicted from the canonicalized feature \(\hat{F}_s\) after Conv2D processing, and its positions are derived from the drive motion \(\alpha_d \) using MLPs, reshaping, and Conv2D operations. Merging both produces a controllable 3D Gaussian for the source portrait.}
    \label{fig:gaussian_generator_and_reenactment_module}
    \vspace{-0.4cm}
\end{figure*}

\textbf{Gaussian Generator and Reenactment Module.} The Gaussian Generator output consists of two parts: the static Gaussian and the dynamic Gaussian. Regarding the static Gaussian, its positions are directly composed of the 3DMM vertices. Other appearance information is obtained by concatenating the global feature and the mesh embedding of the 3DMM vertices and predicting them through multiple layers of MLP. Regarding the dynamic Gaussian, its appearance information is predicted by the canonicalized feature \(\hat{F}_s\) after Conv2D processing. The positions of the dynamic Gaussian are obtained by mapping the drive motion \(\alpha_d \) through MLPs, reshape, Conv2D and other operations. After the static Gaussian and the dynamic Gaussian are merged together, the final result is a model that represents the controllable 3D Gaussian corresponding to the source portrait image. More details can be found in Fig.\ref{fig:gaussian_generator_and_reenactment_module}.
\subsection{D. More Visualization Results}
\begin{figure*}[t]
    \includegraphics[width=1\linewidth]{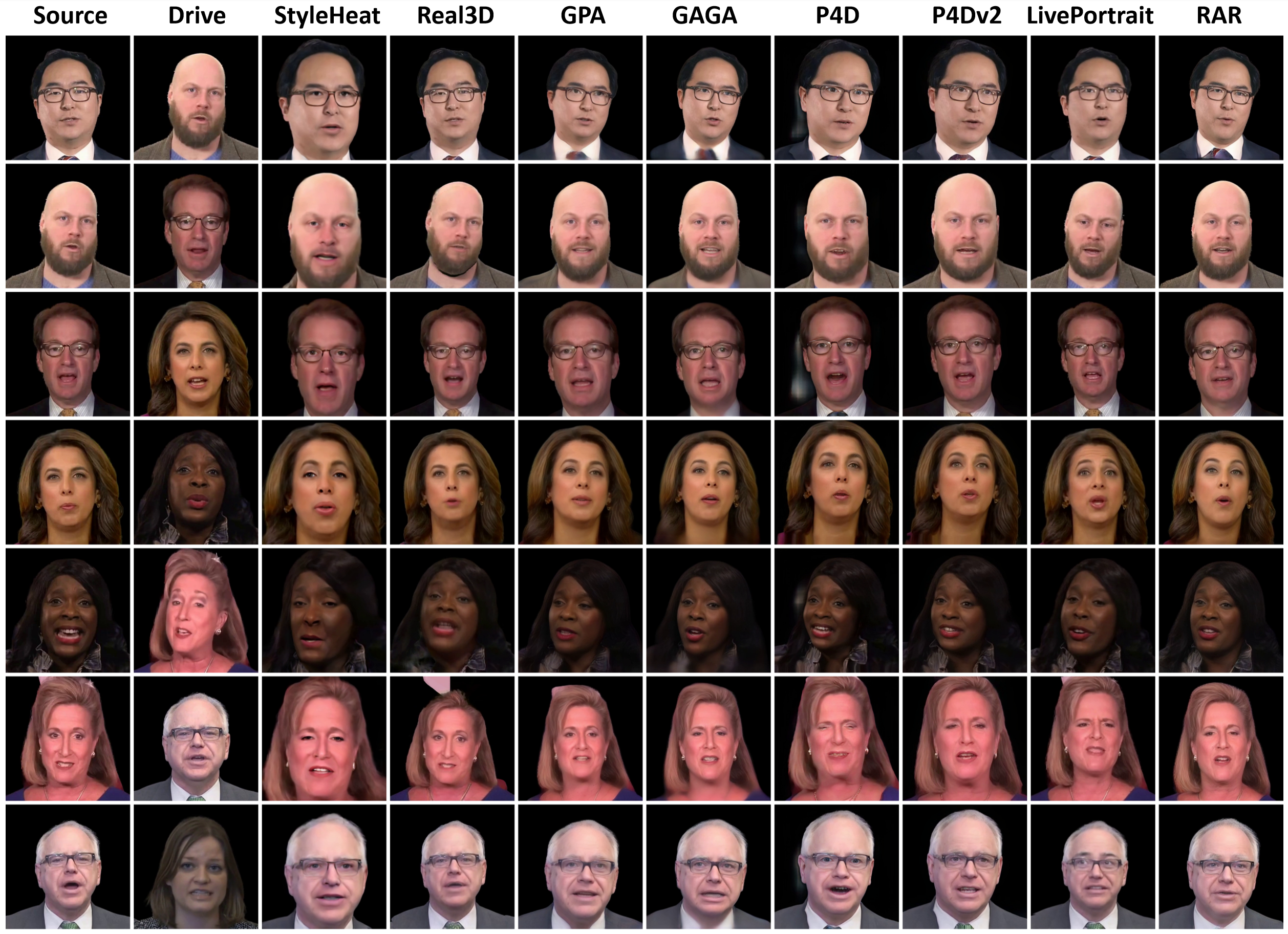}
    
    \caption{visualization of cross-reenacted results on the HDTF dataset.}
    \label{fig:compare_cross_HDTF_supp}
\end{figure*}

\begin{figure*}[t]
    \includegraphics[width=1\linewidth]{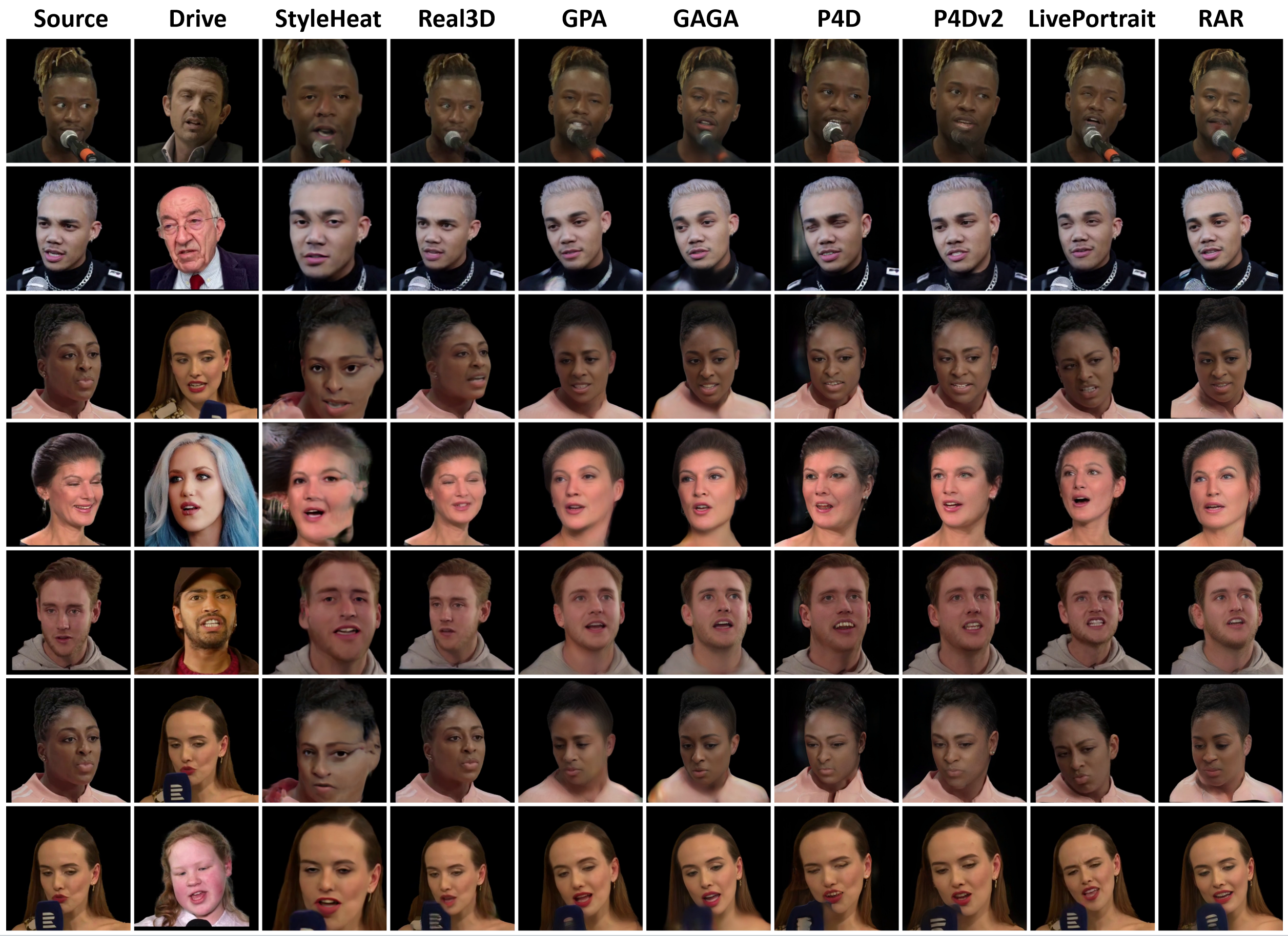}
    
    \caption{visualization of cross-reenacted results on the VFHQ dataset.}
    \label{fig:compare_cross_VFHQ_supp}
\end{figure*}

\begin{figure*}[t]
    \includegraphics[width=1\linewidth]{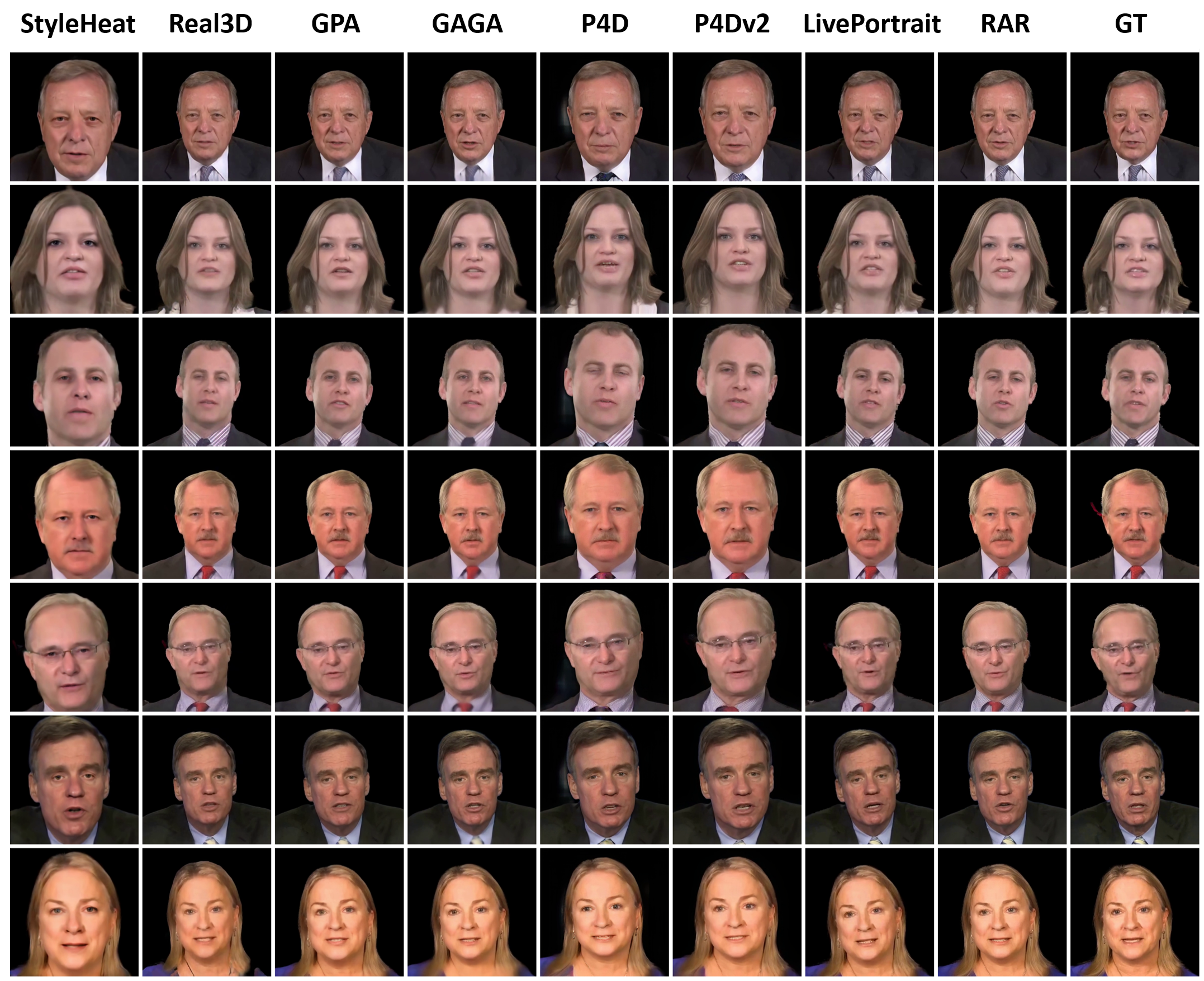}
    
    \caption{visualization of self-reenacted results on the HDTF dataset.}
    \label{fig:compare_self_HDTF_supp}
\end{figure*}

\begin{figure*}[t]
    \includegraphics[width=1\linewidth]{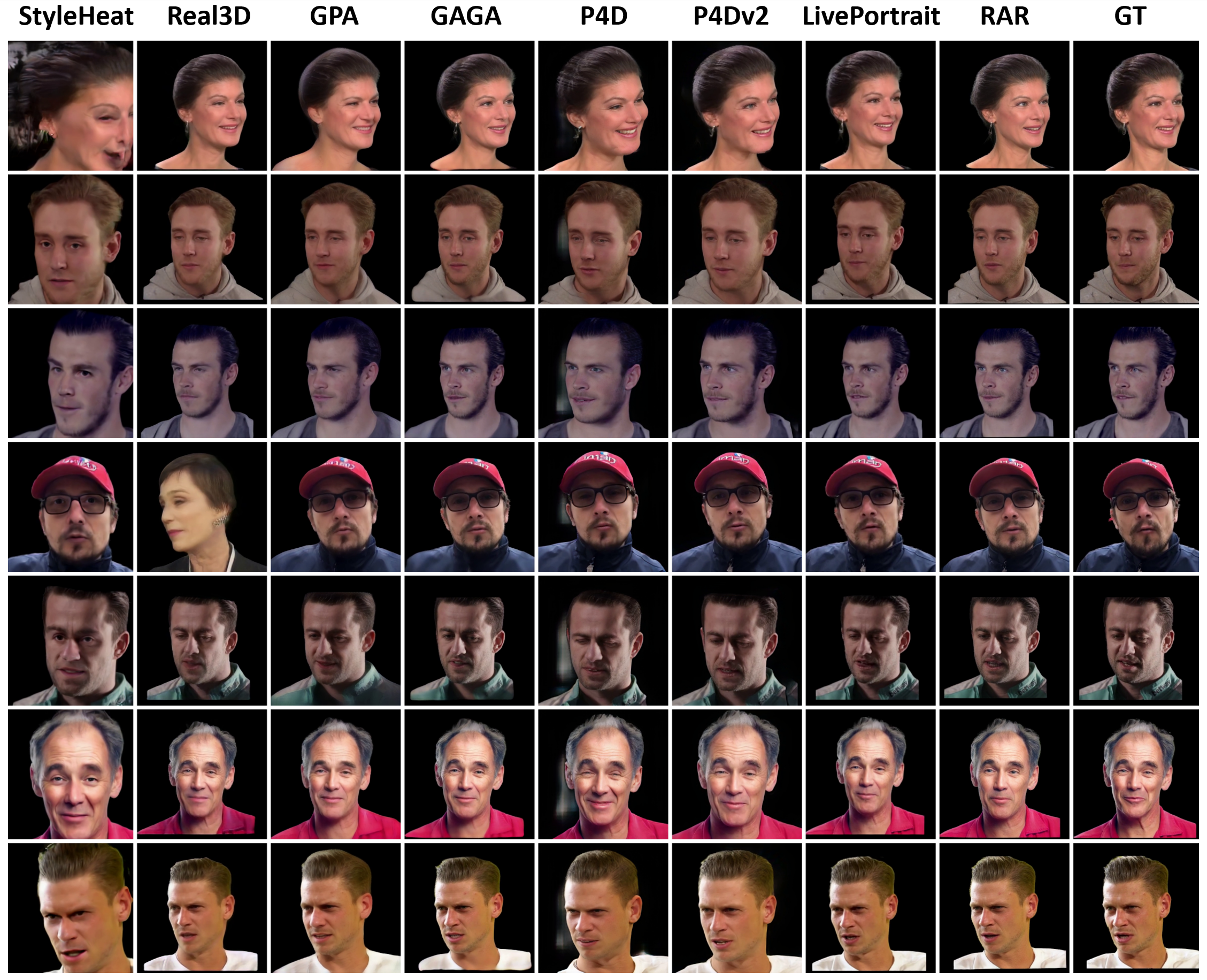}
    
    \caption{visualization of self-reenacted results on the VFHQ dataset.}
    \label{fig:compare_self_VFHQ_supp}
\end{figure*}

In this section, we provide a comprehensive array of additional visualization results that serve to further illustrate our findings. These results include samples from the VFHQ and HDTF test sets, where we apply both self-reenactment and cross-reenactment settings to ensure a thorough examination of our approach under varied conditions. Specifically, by employing a strategic random sampling method, we capture a diverse range of instances that reveal the subtle intricacies in the data, from variations in texture and lighting to the delicate interplay of facial expressions. This process not only highlights the robustness of our technique but also demonstrates its capability to maintain fidelity and realism across different reenactment modalities. You can find all these detailed visual examples, complete with annotations and supplementary insights, in the subsequent Figs.\ref{fig:compare_cross_HDTF_supp},\ref{fig:compare_cross_VFHQ_supp},\ref{fig:compare_self_HDTF_supp},\ref{fig:compare_self_VFHQ_supp}, where each image has been carefully presented to help you gain a deeper understanding of the underlying performance.

\subsection{E. Ethical Consideration}
Our innovative technique can synthesize high-fidelity, real-time talking head video using a single portrait image, leveraging advanced algorithms that capture intricate facial expressions and subtle nuances in movement. By integrating state-of-the-art machine learning models with robust data training over diverse datasets, we ensure that every generated video exudes lifelike clarity and realism. We aspire to extend the application of this pioneering approach to a broad array of endeavors that are not only technically groundbreaking but also significantly advantageous for society—improving remote communications, enhancing digital education, and fostering more inclusive virtual interactions.

As part of our unwavering commitment to ethical advancement in technology, we are enthusiastic about offering dedicated support and practical assistance to the deepfake detection community, which is essential in upholding digital integrity and security. We believe that with careful and judicious application, our method can serve as a catalyst for the wholesome progression of digital human technology by setting new benchmarks in authenticity and efficiency. Ultimately, our work paves the way for transformative innovations across sectors such as entertainment, telehealth, and interactive virtual reality, promising a future where digital and human experiences are more seamlessly integrated.

\end{document}